\pdfoutput=1

\documentclass[11pt]{article}

\usepackage{acl}
\usepackage{float}
\usepackage{times}
\usepackage{latexsym}

\usepackage[T1]{fontenc}

\usepackage[utf8]{inputenc}

\usepackage{microtype}

\usepackage{inconsolata}

\usepackage{bbding,pifont}
\usepackage{graphicx}
\usepackage{comment}
\usepackage{comment}
\usepackage{multirow}
\usepackage{placeins}
\usepackage{xcolor}
\usepackage{amsfonts}
\usepackage{amssymb}
\usepackage{amsmath}
\usepackage{amsthm}
\usepackage{tcolorbox}
\usepackage{subcaption}

\usepackage{array} 
\usepackage{pifont}

\title{
\vspace*{-0.5in}
{{\small \hfill EMNLP-Findings'24}\\
\vspace*{.25in}} 
Insights into LLM Long-Context Failures: \\ When Transformers Know but Don't Tell}

\author{%
  \textbf{Taiming Lu}\textsuperscript{{\ding{80}}} \hspace{0.5cm}
  \textbf{Muhan Gao}\textsuperscript{{\ding{80}}} \hspace{0.5cm}
  \textbf{Kuai Yu}\textsuperscript{{\ding{80}}} \\[0.07cm]  
  \hspace{0.0cm} \textbf{Adam Byerly} \hspace{0.2cm}
  \textbf{Daniel Khashabi} \\[0.12cm]
  \hspace{0.0cm} Johns Hopkins University\\ 
 \hspace{0.0cm} \texttt{\{tlu37, mgao38, kyu25\}@jhu.edu} 
}

\usepackage{soul}

\definecolor{darkred}{RGB}{200,0,0}
\definecolor{lightgreen}{RGB}{231,255,219}
\definecolor{lightred}{RGB}{252,231,234}
\definecolor{lightyellow}{RGB}{250,253,191}
\definecolor{DarkRed}{RGB}{130,25,0}
\definecolor{purplebg}{RGB}{229, 199, 244}

\usepackage{enumitem} 

\newcommand{\tlu}[1]{{\color{purple}[Terry: #1]}}

\definecolor{dcyan}{rgb}{0.2,0.6,0.5}
\newcommand{\adam}[1]{{\color{dcyan}[AB: #1]}}

\begin{document}

\maketitle

\renewcommand{\thefootnote}{\fnsymbol{footnote}}
\footnotetext[0]{{\ding{80}} indicates equal contributions. }

\begin{abstract}
Large Language Models (LLMs) exhibit positional bias, struggling to utilize information from the middle or end of long contexts. Our study explores LLMs' long-context reasoning by probing their hidden representations. We find that while LLMs encode the position of target information, they often fail to leverage this in generating accurate responses. This reveals a disconnect between information \emph{retrieval} and 
its \emph{communication}, a `\textit{know but don't tell}' phenomenon. 
We further analyze task accuracy vs layers,
offering insights into the underlying mechanics of transformer models.
The code for reproducing our results is accessible at \href{https://github.com/TaiMingLu/know-dont-tell}{https://github.com/TaiMingLu/know-dont-tell}.
\end{abstract}



\section{Introduction}
The advent of Transformer-based Large Language Models (LLMs)
has delivered marked improvement in language processing capabilities~\cite{vaswani2017attention,dubey2024llama3herdmodels}. 
These models excel at simultaneously processing extended contexts \cite{ding2024longrope, chen2023extending}, significantly benefiting various downstream tasks like long-text question answering, summarization, and inference~\cite{wang2024novelqa, zhang2024inftybench, shaham-etal-2022-scrolls, shaham-etal-2023-zeroscrolls}.


Despite their advanced capabilities, LLMs often struggle to utilize long inputs fully. This tendency, known as positional bias, leads LLMs to disproportionately prioritize information at the beginning or end of the input sequence~\cite{wang2023primacy}, while crucial details in the middle are frequently overlooked~\cite{liu2023lost}. Numerous strategies have been proposed to address these biases \cite{tang2024middle, li2023split, zhang2024found, goldman2024really}, yet the underlying causes and potential solutions remain unclear. This underscores the need for a deeper investigation into how LLMs handle long-context integration. To fully assess the capabilities of LLMs in handling extended contexts, it is not enough to merely evaluate their final performance: some important information is hidden in models' representations.

\FloatBarrier
\begin{figure}[t!]
\centering
\includegraphics[width=1.0\columnwidth,trim=6.15cm 0.6cm 7.6cm 0.8cm,clip=true]{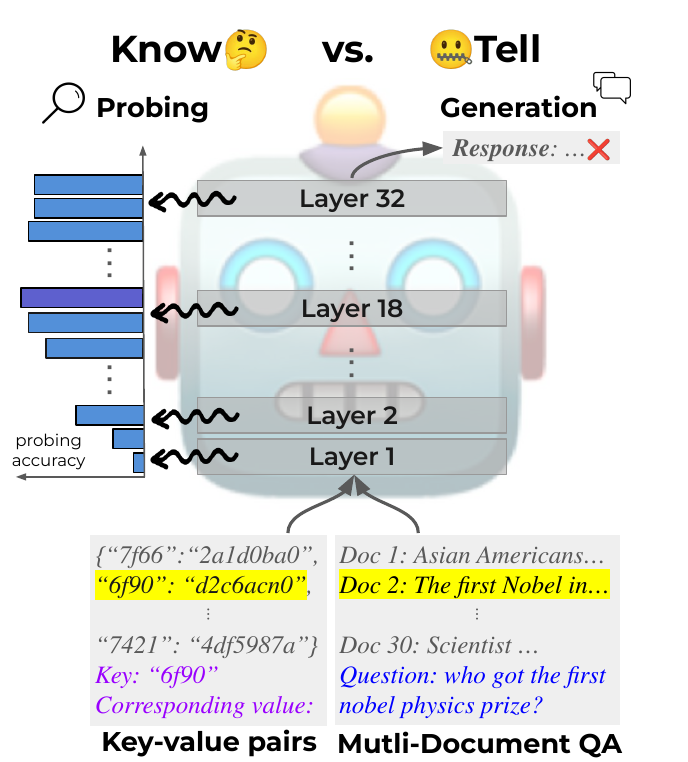}
\caption{Following prompts by \citet{liu2023lost}, we train a probing classifier for each transformer layer to probe the model's ability to identify useful information. The peak accuracy among layers indicates the model's long-context processing effectiveness.}
\label{pipeline}
\end{figure}

This work presents a probing analysis of LLMs' long-context generalization. 
We develop
probes based on LLMs' internal representations for various layers and positions to measure the accuracy of reconstructing the position they correspond to (Fig.~\ref{pipeline}). 
Our working hypothesis is that, \emph{for LLMs to effectively process long contexts, they must encode positional information in their intermediate representations.}

We conduct experiments on two tasks from \citet{liu2023lost} and three recent open-source models. Our findings reveal a gap between the accuracy of LLMs' \emph{generations} and \emph{the probes} on their representations. Notably, while LLM representations identify the position of crucial information within the context (surfaced via probes), they often fail to utilize this information effectively in their responses, leading to what we term the `\textit{Know but don't Tell}' phenomenon. To our knowledge, this is the first work to use probing analysis to highlight this observation. We hope that our work on distinguishing ``knowing'' and ``telling'' motivates future work on tackling LLMs' long-context challenges.


In summary, our contributions are as follows:
(1) \underline{Probing analysis:} We introduce a novel framework to investigate the long-context reasoning capabilities of LLMs. This framework allows us to measure how accurately LLMs encode positional information across various layers and positions within their intermediate representations.
(2) \underline{Empirical evaluation:} We conduct comprehensive experiments using tasks from \citet{liu2023lost} and three recent open-source models. Our empirical evaluation provides new insights into the positional biases of LLMs and their impact on model performance.
(3) \underline{\emph{`Know but Don't Tell'} phenomenon:} Our analysis reveals a critical gap between LLMs' ability to encode and utilize positional information. We identify the ``\textit{Know but don't Tell}'' phenomenon, where LLMs accurately identify the position of crucial information but fail to leverage this knowledge in generating accurate responses.
By distinguishing between the encoding and utilization of positional information, our work lays the foundation for future advancements in LLM performance and reliability.

\section{Related Work}

\paragraph{Positional bias.} 
LLMs exhibit a positional bias, where the location of crucial context information influences their performance \cite{zhao2021calibrate}. One prominent example is the ``lost in the middle'' phenomenon, where comprehension declines for information in the center of a long context \cite{liu2023lost}. Additionally, recency bias is observed, particularly in few-shot learning scenarios, where models tend to favor information near the end of the prompt \cite{zhao2021calibrate}. Such biases could stem from the positioning of key data in pre-training sets, which often places important elements near critical points \cite{peysakhovich2023attention}. Our work delves into this phenomenon by examining the underlying mechanisms within the transformer layers of LLMs.

\paragraph{Probing.}
Probing classifiers are extensively used to elucidate the inner workings of LLMs~\cite{alain2016understanding, azaria2023internal, jin2024exploring, ju2024large, scaling, jiang2024llms, sky2024androids}. 
Various works train probes on model representations 
to assess how well they encode various
linguistic features, such as phrase-level, syntactic, and semantic information~\cite{liu2023cognitive, marks2023geometry, li2024inference}. The efficacy of a classifier in a given task indicates the degree to which that layer successfully captures pertinent information. Our study employs probing as a proxy to determine whether the LLMs accurately identify and represent crucial parts of the context.

\section{Experimental Setup}
\label{sec:probing}

We conduct a layer-wise probing analysis to determine if the model accurately identifies target information from a given prompt. We expect that higher probing accuracy shows a stronger connection between the model's hidden representations and its internal understanding of the target information.

\paragraph{Datasets and prompts.}
We follow the datasets and prompts used by \citet{liu2023lost}. Our datasets include: 
(1) \underline{Key-Value Pairs Retrieval} (kv-pairs)  where the context contains a collection of keys and their corresponding values (128-bit randomly generated UUIDs). The goal of this task is to identify a value given its key. Each prompt for this task contains 100 kv-pairs and a target key.  
(2) \underline{Multi-Document Question Answering} (MDQA) where the context contains multiple sets of evidence paragraphs. The goal of this task is to, given a question, identify the relevant document and produce an answer. Each prompt for this task contains 30 documents, and a target question. Given a set of key-value pairs/documents, with only one containing target information, the LLM is prompted to output the value/answer given the key/question. Details (e.g., prompts) are in \S\ref{sec:prompt}.


\begin{figure*}[!htb]
  \centering
  \begin{subfigure}{\textwidth}
    \centering
    \caption{LLaMa3-8B-Instruct}
    \includegraphics[width=1\linewidth]{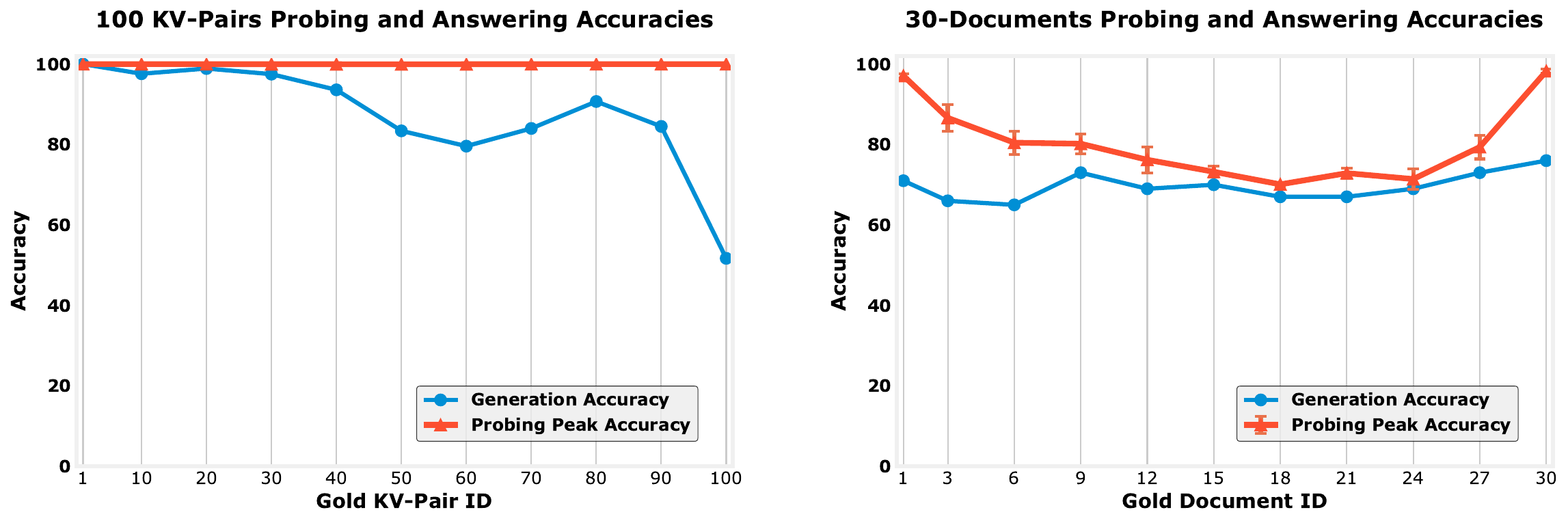}
    \label{subfig:probing_accuracy}

  \end{subfigure}
  
  \vspace{0.2cm}
  
  \begin{subfigure}{\textwidth}
    \hspace{3mm}
    \centering
    \caption{Mistral-7B-Instruct-v0.3}
    \includegraphics[width=1\linewidth]{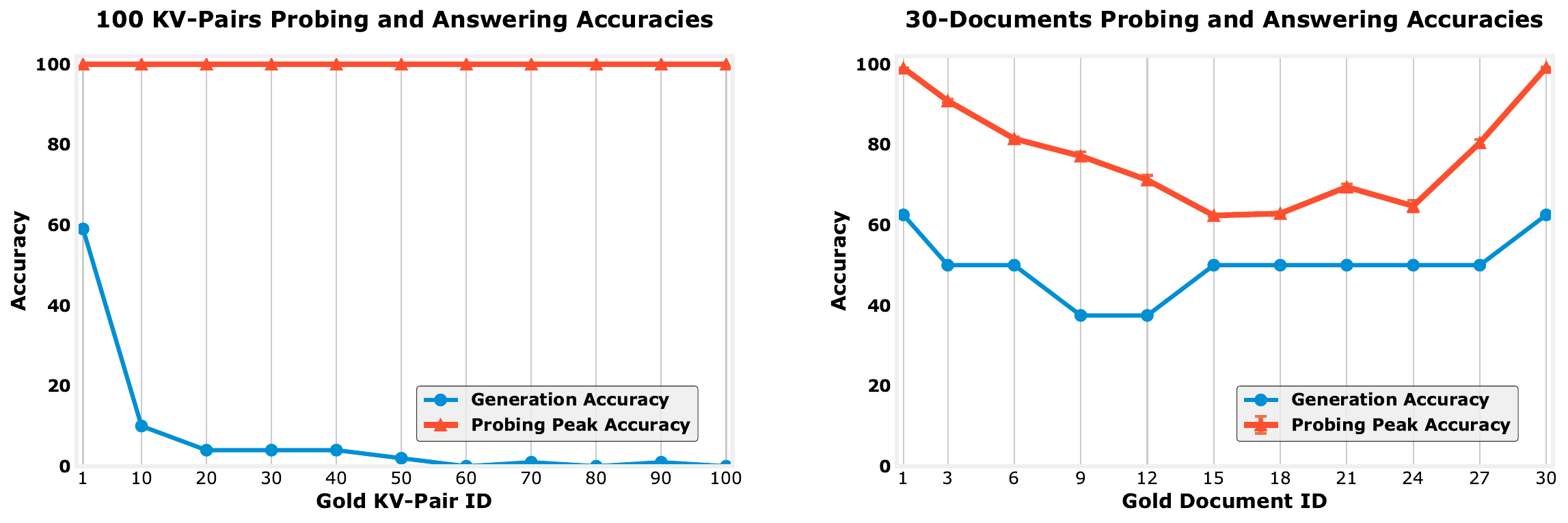}
    
    \label{subfig:mistral}
  \end{subfigure}
  
  \caption{
  Accuracy of LLMs in directly generating answers (blue line) compared to the maximum probing accuracy across layers by our probing classifiers (red line). In both tasks, our probing classifiers surpass the model's generated answers across all gold positions. \textbf{This highlights a distinction between \emph{knowing} the context and \emph{utilizing} it.}
    }
  \label{fig:probing_layer_section_3}
\end{figure*}

\paragraph{Probing classifiers.}
For each input prompt, we extract the last token embedding $x^l$ from \emph{each layer} $l$. We then train a separate linear classifier for each layer $l$, characterized by weight $\mathbf{w}_l$ and bias $b_l$. The classifier takes $x^l$ as input to predict the gold kv-pair or document ID (position among all pairs/documents). The classifier aims to minimize the following objective:

\vspace{-0.1cm}
$$
J(\mathbf{\Theta}) = -\frac{1}{N} \sum_{i=1}^{N} \log \left( \frac{e^{w_{y_i}^\top x_i^l + b_{y_i}}}{\sum_{j=1}^{C} e^{W_j^\top x_j^l + b_{j}}} \right),
$$
where $\mathbf{\Theta} = \{ \mathbf{w}_l, b_l \}_{l=1}^L$ is the union of all parameters,
 \(L\) is the number of layers,
 \(N\) is the number of data points, 
 \(C\) is the number of different gold IDs (classes), \(x_i^l\) represents the input embedding of the $l$-th layer from the \(i\)-th data point, \(y_{i}\) is the label for the \(i\)-th data point.
Ultimately, this recipe gives one probing classifier for embeddings of \emph{each layer}. Using these models, we show results per layer and across layers. We select training and test datasets that are \textit{mutually exclusive} across each other, containing entirely different sets of key-value pairs/documents. Thus, neither classifiers nor the language model had access to the test samples' gold pair/document ID during the evaluation. We provide implementation details in~\S\ref{sec:dataset}.

\begin{figure*}[tbh]
  \centering
  \begin{subfigure}{\textwidth}
    \centering
    \caption{LLaMa3-8B-Instruct}
    \includegraphics[width=1\linewidth]{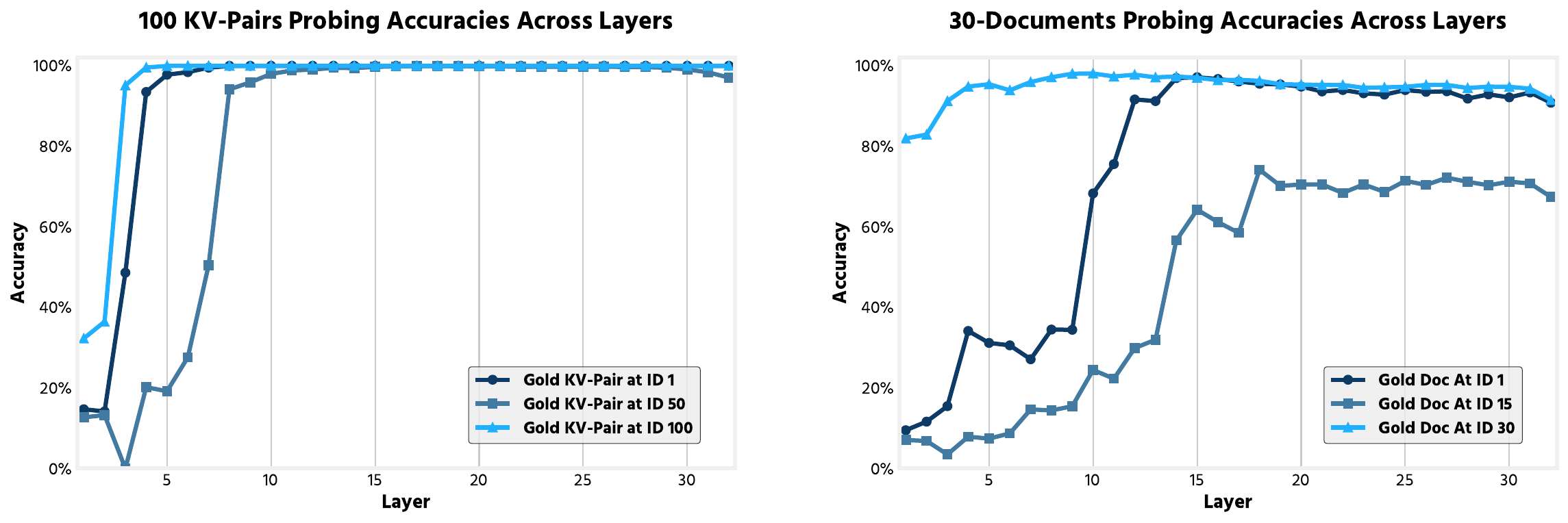}
    \label{subfig:probing_layer}
  \end{subfigure}
  \vspace{0.2cm}
  \begin{subfigure}{\textwidth}
    \centering
    \caption{Mistral-7B-Instruct-v0.3}
    \includegraphics[width=1\linewidth]{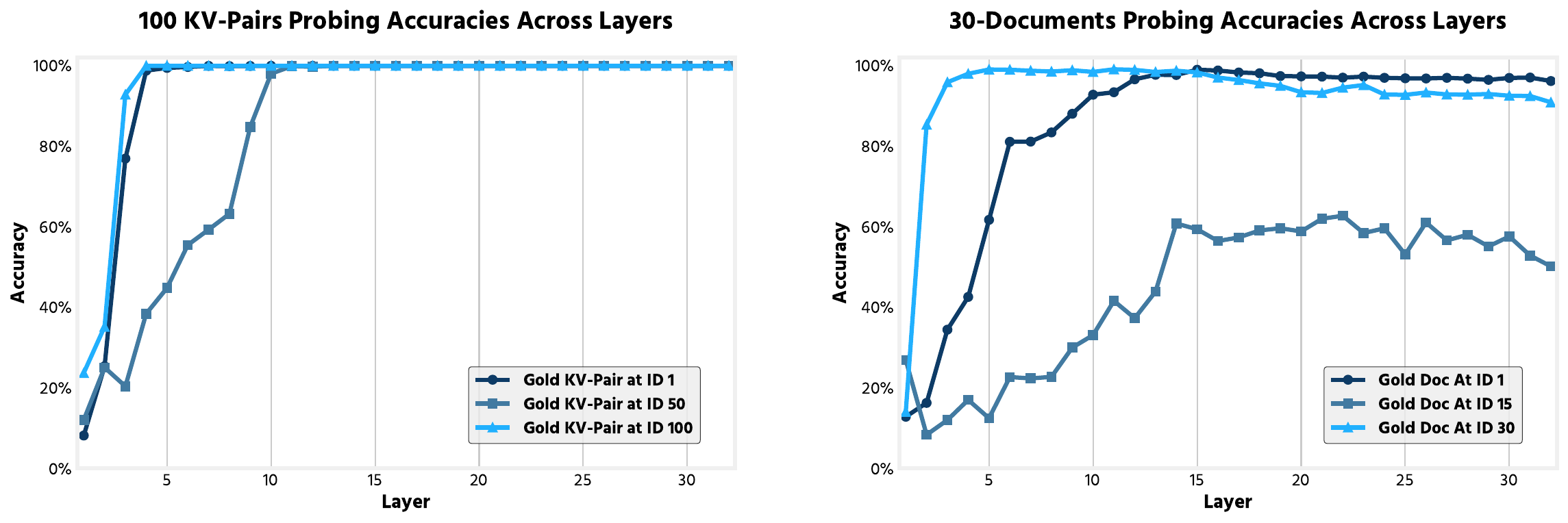}
    \label{subfig:mistral_probing}
  \end{subfigure}
  \caption{
   The probing accuracy for each layer in the two tasks: kv-pairs (left) and MDQA (right). Different colors represent the position of target information within the input context. In both tasks, \textbf{extracting mid-context information requires more layers.}
    }
  \label{fig:probing_comparison_section_4}
\end{figure*}

\paragraph{Models and hyperparameters.}
We employ two SOTA models, \texttt{LLaMa3-8B-Instruct}~\cite{llama3modelcard} and \texttt{
Mistral-7B-Instruct-v0.3}~\cite{jiang2023mistral} to do our probing analysis. 
The results for the third model, \texttt{Gemma}~\cite{team2024gemma}, are provided in the Appendix due to space constraints.
To minimize uncertainty from random initialization, each classifier is trained ten times. We report the mean and standard deviation (error bars) from these ten independent experiments. 
As we observe in \S\ref{sec:results}, consistency across the three models confirms that the \emph{`Know but Don’t Tell'} phenomenon is not specific to a single model's architecture.


\paragraph{Metrics.}
Following~\citet{liu2023lost}, we use \emph{accuracy} to evaluate our models' success. We measure two types of accuracy: (a) \underline{Generation} accuracy, which quantifies how well LLMs generate the correct value in kv-pair retrieval or the correct answer string in MDQA, and (b) \underline{Probing} accuracy, which evaluates how accurately classifiers predict the gold kv-pair or document ID, quantifying the extent to which layers encode information from the input context.

\section{LLMs Know but Don't Tell}
\label{sec:results}

\subsection{Experiment: Peak Probing Accuracy Across LLM Layers}
\label{sec:4.1}

We examine the maximum accuracy across all transformer layers as an indicator of whether the model successfully identifies key information within the prompt during the forward pass. Specifically, we choose the probing classifier with the highest accuracy among all layers. This peak layer probing accuracy is displayed in Fig.~\ref{fig:probing_layer_section_3}. For comparison, we also present the accuracy of LLMs in generating the answer, independent of our probing classifiers.


\paragraph{LLMs know but don't tell.}
Our results indicate that the model's hidden representations \emph{do} contain information about the location of the target information. Specifically, in the kv-pairs setup (Fig.~\ref{fig:probing_layer_section_3}; left) there is always a layer where its probe can almost perfectly identify the location of the correct key-value pair associated with the prompt. 
 This holds true even when the LLM provides an incorrect answer or suggests no gold information is present in the input.
This suggests a disconnect between the model's ability to locate the information and generate a response based on that information.

A similar trend is also observed for MDQA (Fig.~\ref{fig:probing_layer_section_3}; right) where the peak probing accuracy is consistently higher than the direct answer accuracy, indicating the same disconnect. These findings highlight that while the model can recognize and encode the location of relevant information within its layers, this knowledge does not always translate into an accurate generation answer. 


\subsection{Experiment: Probing Across Layers}
\label{sec:4:2}
We focus on the probing classifiers' accuracy across LLM layers to understand the flow of information across an LLM's layers. Fig.~\ref{fig:probing_comparison_section_4} visualizes the probing classifier accuracy per layer. In both kv-pairs and MDQA setups, we show this accuracy for three positions: 
when the target information is at the start, middle, or end of the input.

\paragraph{Middle-context information requires more layers to be located.}
Our results reveal that LLMs locate target information gradually at early layers. Specifically, in the kv-pair setup (Fig.~\ref{fig:probing_comparison_section_4}; left), probing accuracy consistently increases until it reaches perfect accuracy at layer 13. Notably, when the target kv-pair is at the middle position of the input prompt, LLM requires more layers to locate the target information.

The general trends of the MDQA scenario (Fig.~\ref{fig:probing_comparison_section_4}; right) are similar in principle but with nuanced differences. The patterns vary significantly with the position of target information. Classifiers perform best when the target document is at the start or end of the input context, with early layers near-perfect prediction maintaining across subsequent layers.
 However, it takes more layers for the probing classifier to achieve peak accuracy when the gold information is in the middle or tail of the context. Interestingly, classifier accuracy decreases after the peak when the target document is in the middle. 
 
Related to the results in this section, in \S\ref{sec:visualization} we present visualizations of LLM intermediate representations using PCA and UMAP dimension reduction. In these visualizations, projected representations corresponding to adjacent gold documents are connected by lines. We observe that as the number of layers increases, particularly in the middle layers, the projected representations are spaced apart and arranged in a path. In contrast, these embeddings are more entangled in the earlier layers. This further corroborates our probing results.

\begin{figure}[thb!]
  \includegraphics[width=\columnwidth,trim=0.35cm 0cm 0.2cm 0cm]{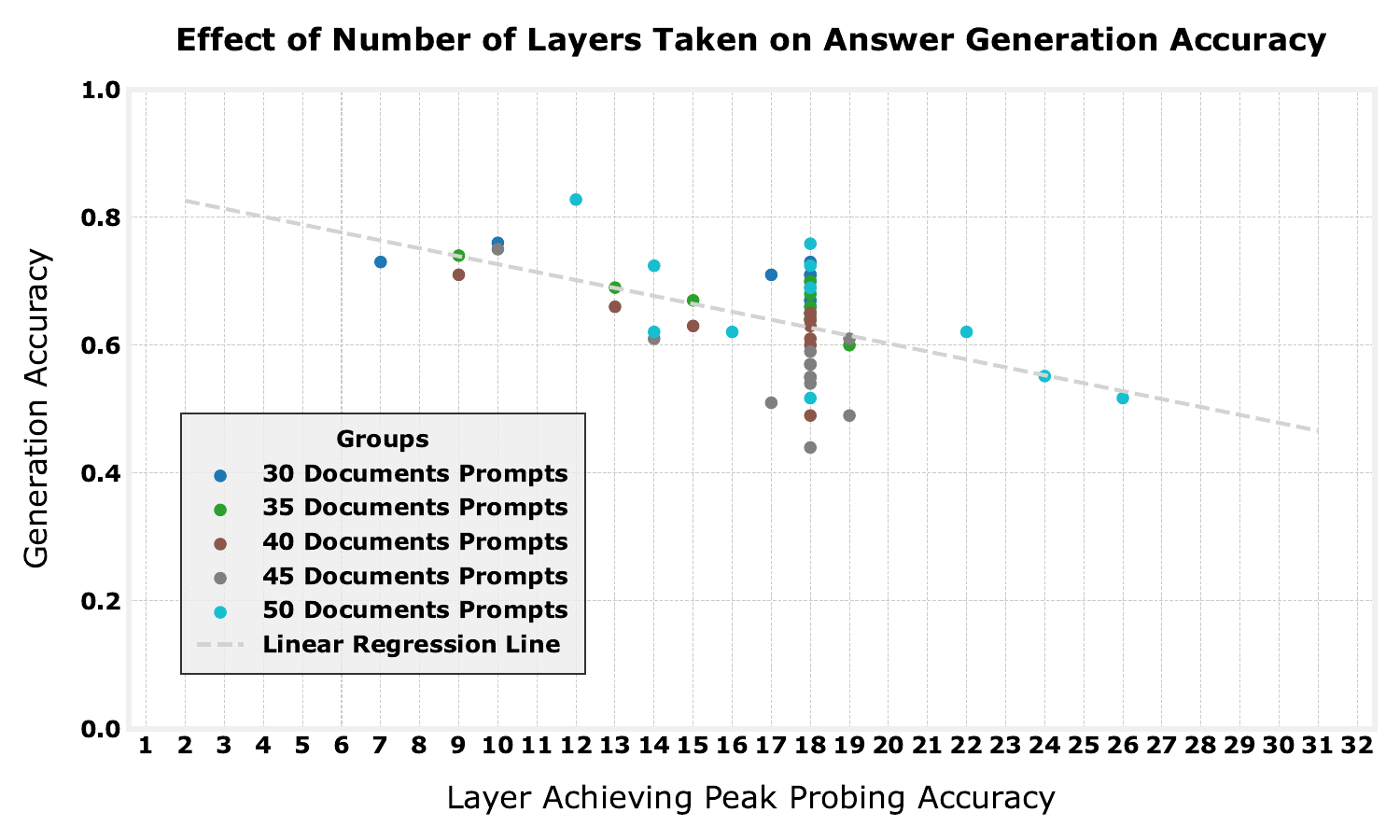}
  \caption{
  The LLM layer that achieves the peak probing accuracy ($x$-axis) vs. the accuracy of LLM in generating the correct answer  ($y$-axis).
  We observe that a \emph{later} peak correlates with \emph{lower} accuracy in the language model's final output. 
  This implies that 
  the earlier an LLM encodes information from a specific index, the higher the accuracy of the final output for that position. 
  }
  \label{fig:peak-layer}
\end{figure}

\subsection{Experiment: Number of Layers Taken for Locating Target Information}
\label{sec:extraction timing}
Our probing experiments (\S\ref{sec:4:2}) reveal that the model's encoding of target information position improves with more layers but then it degrades. 
This motivates, the investigation of 
the relationship between the number of layers needed by the model to locate 
 target information from the prompt and the LLM's  accuracy in generating that target information.

We conduct additional multi-document probing tasks with 35, 40, 45, and 50 documents. 
 In Fig.~\ref{fig:peak-layer}, for all IDs with probing accuracy exceeding 60\% (to minimize the impact of outliers), the $x$-axis represents the layer where peak probing accuracy is achieved, while the $y$-axis displays the LLM's generation accuracy (without involving probes).

\paragraph{Early-layer information localization leads to higher generation accuracy.}
As Fig.~\ref{fig:peak-layer} shows, there is a negative correlation between the layer with peak accuracy of locating target information and its final output accuracy.
A two-sided t-test was employed to confirm the statistical significance of this observation (p-value < 5e-5; null hypothesis: zero slope).
This negative correlation implies that the earlier the model identifies the target document within its layers, the more likely it is to generate an accurate final answer.

\section{Conclusion}
Our study shows that LLMs capture context internally but fail to generate the correct answer due to positional bias. Probing experiments reveal that middle-context information requires deeper layers to be accessed, reducing performance. We also found that delayed extraction leads to lower accuracy. We hope these findings shed more light on the underlying cause behind positional bias and point to new areas for improving long-range information processing in LLMs.

\newpage

\section{Limitation}


Knowledge of the gold document's location and the ability to cite from it are distinct but connected; the model might know the location but still, fail to integrate it into a coherent and accurate answer. This comparison does not fully capture the nuanced interactions between the model's internal attention mechanisms and output generation capabilities. While these limitations are acknowledged, they do not detract from the core contributions of our work. Our findings provide insights into the positional effects on model performance and highlight the importance of document sequence in information retrieval tasks. By identifying specific areas where the model struggles, we lay the groundwork for future improvements and optimizations in model design and training.


\section*{Acknowledgements}
This work is supported by a generous gift the Allen Institute for AI and partly by ONR grant (N00014-24-1-2089). 
We are grateful to the anonymous reviewers for their constructive feedback. 
GPU machines for conducting experiments were provided by ARCH Rockfish cluster.

\bibliography{custom,ref}

\providecommand{\CNFX}[1]{{\em{\textrm{(#1)}}}}
\begin{thebibliography}{32}
\providecommand{\natexlab}[1]{#1}

\bibitem[{AI@Meta(2024)}]{llama3modelcard}
AI@Meta. 2024.
\newblock \href {https://github.com/meta-llama/llama3/blob/main/MODEL_CARD.md} {Llama 3 model card}.

\bibitem[{Alain and Bengio(2016)}]{alain2016understanding}
Guillaume Alain and Yoshua Bengio. 2016.
\newblock Understanding intermediate layers using linear classifier probes.
\newblock \emph{arXiv preprint arXiv:1610.01644}.

\bibitem[{Azaria and Mitchell(2023)}]{azaria2023internal}
Amos Azaria and Tom Mitchell. 2023.
\newblock The internal state of an llm knows when its lying.
\newblock \emph{arXiv preprint arXiv:2304.13734}.

\bibitem[{Chen et~al.(2023)Chen, Wong, Chen, and Tian}]{chen2023extending}
Shouyuan Chen, Sherman Wong, Liangjian Chen, and Yuandong Tian. 2023.
\newblock \href {https://arxiv.org/abs/2306.15595} {Extending context window of large language models via positional interpolation}.
\newblock \emph{arXiv preprint arXiv:2306.15595}.

\bibitem[{Ding et~al.(2024)Ding, Zhang, Zhang, Xu, Shang, Xu, Yang, and Yang}]{ding2024longrope}
Yiran Ding, Li~Lyna Zhang, Chengruidong Zhang, Yuanyuan Xu, Ning Shang, Jiahang Xu, Fan Yang, and Mao Yang. 2024.
\newblock \href {https://arxiv.org/abs/2402.13753} {Longrope: Extending llm context window beyond 2 million tokens}.
\newblock \emph{Preprint}, arXiv:2402.13753.

\bibitem[{Dubey et~al.(2024)Dubey, Jauhri, Pandey, Kadian, Al-Dahle, Letman, Mathur, Schelten, Yang, Fan, Goyal, Hartshorn, Yang, Mitra, Sravankumar, Korenev, Hinsvark, Rao, Zhang, Rodriguez, Gregerson, Spataru, Roziere, Biron, Tang, Chern, Caucheteux, Nayak, Bi, Marra, McConnell, Keller, Touret, Wu, Wong, Ferrer, Nikolaidis, Allonsius, Song, Pintz, Livshits, Esiobu, Choudhary, Mahajan, Garcia-Olano, Perino, Hupkes, Lakomkin, AlBadawy, Lobanova, Dinan, Smith, Radenovic, Zhang, Synnaeve, Lee, Anderson, Nail, Mialon, Pang, Cucurell, Nguyen, Korevaar, Xu, Touvron, Zarov, Ibarra, Kloumann, Misra, Evtimov, Copet, Lee, Geffert, Vranes, Park, Mahadeokar, Shah, van~der Linde, Billock, Hong, Lee, Fu, Chi, Huang, Liu, Wang, Yu, Bitton, Spisak, Park, Rocca, Johnstun, Saxe, Jia, Alwala, Upasani, Plawiak, Li, Heafield, Stone, El-Arini, Iyer, Malik, Chiu, Bhalla, Rantala-Yeary, van~der Maaten, Chen, Tan, Jenkins, Martin, Madaan, Malo, Blecher, Landzaat, de~Oliveira, Muzzi, Pasupuleti, Singh, Paluri, Kardas, Oldham, Rita,
  Pavlova, Kambadur, Lewis, Si, Singh, Hassan, Goyal, Torabi, Bashlykov, Bogoychev, Chatterji, Duchenne, \c{C}elebi, Alrassy, Zhang, Li, Vasic, Weng, Bhargava, Dubal, Krishnan, Koura, Xu, He, Dong, Srinivasan, Ganapathy, Calderer, Cabral, Stojnic, Raileanu, Girdhar, Patel, Sauvestre, Polidoro, Sumbaly, Taylor, Silva, Hou, Wang, Hosseini, Chennabasappa, Singh, Bell, Kim, Edunov, Nie, Narang, Raparthy, Shen, Wan, Bhosale, Zhang, Vandenhende, Batra, Whitman, Sootla, Collot, Gururangan, Borodinsky, Herman, Fowler, Sheasha, Georgiou, Scialom, Speckbacher, Mihaylov, Xiao, Karn, Goswami, Gupta, Ramanathan, Kerkez, Gonguet, Do, Vogeti, Petrovic, Chu, Xiong, Fu, Meers, Martinet, Wang, Tan, Xie, Jia, Wang, Goldschlag, Gaur, Babaei, Wen, Song, Zhang, Li, Mao, Coudert, Yan, Chen, Papakipos, Singh, Grattafiori, Jain, Kelsey, Shajnfeld, Gangidi, Victoria, Goldstand, Menon, Sharma, Boesenberg, Vaughan, Baevski, Feinstein, Kallet, Sangani, Yunus, Lupu, Alvarado, Caples, Gu, Ho, Poulton, Ryan, Ramchandani, Franco, Saraf,
  Chowdhury, Gabriel, Bharambe, Eisenman, Yazdan, James, Maurer, Leonhardi, Huang, Loyd, Paola, Paranjape, Liu, Wu, Ni, Hancock, Wasti, Spence, Stojkovic, Gamido, Montalvo, Parker, Burton, Mejia, Wang, Kim, Zhou, Hu, Chu, Cai, Tindal, Feichtenhofer, Civin, Beaty, Kreymer, Li, Wyatt, Adkins, Xu, Testuggine, David, Parikh, Liskovich, Foss, Wang, Le, Holland, Dowling, Jamil, Montgomery, Presani, Hahn, Wood, Brinkman, Arcaute, Dunbar, Smothers, Sun, Kreuk, Tian, Ozgenel, Caggioni, Guzm\'{a}n, Kanayet, Seide, Florez, Schwarz, Badeer, Swee, Halpern, Thattai, Herman, Sizov, Guangyi, Zhang, Lakshminarayanan, Shojanazeri, Zou, Wang, Zha, Habeeb, Rudolph, Suk, Aspegren, Goldman, Damlaj, Molybog, Tufanov, Veliche, Gat, Weissman, Geboski, Kohli, Asher, Gaya, Marcus, Tang, Chan, Zhen, Reizenstein, Teboul, Zhong, Jin, Yang, Cummings, Carvill, Shepard, McPhie, Torres, Ginsburg, Wang, Wu, U, Saxena, Prasad, Khandelwal, Zand, Matosich, Veeraraghavan, Michelena, Li, Huang, Chawla, Lakhotia, Huang, Chen, Garg, A, Silva, Bell,
  Zhang, Guo, Yu, Moshkovich, Wehrstedt, Khabsa, Avalani, Bhatt, Tsimpoukelli, Mankus, Hasson, Lennie, Reso, Groshev, Naumov, Lathi, Keneally, Seltzer, Valko, Restrepo, Patel, Vyatskov, Samvelyan, Clark, Macey, Wang, Hermoso, Metanat, Rastegari, Bansal, Santhanam, Parks, White, Bawa, Singhal, Egebo, Usunier, Laptev, Dong, Zhang, Cheng, Chernoguz, Hart, Salpekar, Kalinli, Kent, Parekh, Saab, Balaji, Rittner, Bontrager, Roux, Dollar, Zvyagina, Ratanchandani, Yuvraj, Liang, Alao, Rodriguez, Ayub, Murthy, Nayani, Mitra, Li, Hogan, Battey, Wang, Maheswari, Howes, Rinott, Bondu, Datta, Chugh, Hunt, Dhillon, Sidorov, Pan, Verma, Yamamoto, Ramaswamy, Lindsay, Lindsay, Feng, Lin, Zha, Shankar, Zhang, Zhang, Wang, Agarwal, Sajuyigbe, Chintala, Max, Chen, Kehoe, Satterfield, Govindaprasad, Gupta, Cho, Virk, Subramanian, Choudhury, Goldman, Remez, Glaser, Best, Kohler, Robinson, Li, Zhang, Matthews, Chou, Shaked, Vontimitta, Ajayi, Montanez, Mohan, Kumar, Mangla, Albiero, Ionescu, Poenaru, Mihailescu, Ivanov, Li, Wang,
  Jiang, Bouaziz, Constable, Tang, Wang, Wu, Wang, Xia, Wu, Gao, Chen, Hu, Jia, Qi, Li, Zhang, Zhang, Adi, Nam, Yu, Wang, Hao, Qian, He, Rait, DeVito, Rosnbrick, Wen, Yang, and Zhao}]{dubey2024llama3herdmodels}
Abhimanyu Dubey, Abhinav Jauhri, Abhinav Pandey, Abhishek Kadian, Ahmad Al-Dahle, Aiesha Letman, Akhil Mathur, Alan Schelten, Amy Yang, Angela Fan, Anirudh Goyal, Anthony Hartshorn, Aobo Yang, Archi Mitra, Archie Sravankumar, Artem Korenev, Arthur Hinsvark, Arun Rao, Aston Zhang, Aurelien Rodriguez, Austen Gregerson, Ava Spataru, Baptiste Roziere, Bethany Biron, Binh Tang, Bobbie Chern, Charlotte Caucheteux, Chaya Nayak, Chloe Bi, Chris Marra, Chris McConnell, Christian Keller, Christophe Touret, Chunyang Wu, Corinne Wong, Cristian~Canton Ferrer, Cyrus Nikolaidis, Damien Allonsius, Daniel Song, Danielle Pintz, Danny Livshits, David Esiobu, Dhruv Choudhary, Dhruv Mahajan, Diego Garcia-Olano, Diego Perino, Dieuwke Hupkes, Egor Lakomkin, Ehab AlBadawy, Elina Lobanova, Emily Dinan, Eric~Michael Smith, Filip Radenovic, Frank Zhang, Gabriel Synnaeve, Gabrielle Lee, Georgia~Lewis Anderson, Graeme Nail, Gregoire Mialon, Guan Pang, Guillem Cucurell, Hailey Nguyen, Hannah Korevaar, Hu~Xu, Hugo Touvron, Iliyan Zarov,
  Imanol~Arrieta Ibarra, Isabel Kloumann, Ishan Misra, Ivan Evtimov, Jade Copet, Jaewon Lee, Jan Geffert, Jana Vranes, Jason Park, Jay Mahadeokar, Jeet Shah, Jelmer van~der Linde, Jennifer Billock, Jenny Hong, Jenya Lee, Jeremy Fu, Jianfeng Chi, Jianyu Huang, Jiawen Liu, Jie Wang, Jiecao Yu, Joanna Bitton, Joe Spisak, Jongsoo Park, Joseph Rocca, Joshua Johnstun, Joshua Saxe, Junteng Jia, Kalyan~Vasuden Alwala, Kartikeya Upasani, Kate Plawiak, Ke~Li, Kenneth Heafield, Kevin Stone, Khalid El-Arini, Krithika Iyer, Kshitiz Malik, Kuenley Chiu, Kunal Bhalla, Lauren Rantala-Yeary, Laurens van~der Maaten, Lawrence Chen, Liang Tan, Liz Jenkins, Louis Martin, Lovish Madaan, Lubo Malo, Lukas Blecher, Lukas Landzaat, Luke de~Oliveira, Madeline Muzzi, Mahesh Pasupuleti, Mannat Singh, Manohar Paluri, Marcin Kardas, Mathew Oldham, Mathieu Rita, Maya Pavlova, Melanie Kambadur, Mike Lewis, Min Si, Mitesh~Kumar Singh, Mona Hassan, Naman Goyal, Narjes Torabi, Nikolay Bashlykov, Nikolay Bogoychev, Niladri Chatterji, Olivier
  Duchenne, Onur \c{C}elebi, Patrick Alrassy, Pengchuan Zhang, Pengwei Li, Petar Vasic, Peter Weng, Prajjwal Bhargava, Pratik Dubal, Praveen Krishnan, Punit~Singh Koura, Puxin Xu, Qing He, Qingxiao Dong, Ragavan Srinivasan, Raj Ganapathy, Ramon Calderer, Ricardo~Silveira Cabral, Robert Stojnic, Roberta Raileanu, Rohit Girdhar, Rohit Patel, Romain Sauvestre, Ronnie Polidoro, Roshan Sumbaly, Ross Taylor, Ruan Silva, Rui Hou, Rui Wang, Saghar Hosseini, Sahana Chennabasappa, Sanjay Singh, Sean Bell, Seohyun~Sonia Kim, Sergey Edunov, Shaoliang Nie, Sharan Narang, Sharath Raparthy, Sheng Shen, Shengye Wan, Shruti Bhosale, Shun Zhang, Simon Vandenhende, Soumya Batra, Spencer Whitman, Sten Sootla, Stephane Collot, Suchin Gururangan, Sydney Borodinsky, Tamar Herman, Tara Fowler, Tarek Sheasha, Thomas Georgiou, Thomas Scialom, Tobias Speckbacher, Todor Mihaylov, Tong Xiao, Ujjwal Karn, Vedanuj Goswami, Vibhor Gupta, Vignesh Ramanathan, Viktor Kerkez, Vincent Gonguet, Virginie Do, Vish Vogeti, Vladan Petrovic, Weiwei
  Chu, Wenhan Xiong, Wenyin Fu, Whitney Meers, Xavier Martinet, Xiaodong Wang, Xiaoqing~Ellen Tan, Xinfeng Xie, Xuchao Jia, Xuewei Wang, Yaelle Goldschlag, Yashesh Gaur, Yasmine Babaei, Yi~Wen, Yiwen Song, Yuchen Zhang, Yue Li, Yuning Mao, Zacharie~Delpierre Coudert, Zheng Yan, Zhengxing Chen, Zoe Papakipos, Aaditya Singh, Aaron Grattafiori, Abha Jain, Adam Kelsey, Adam Shajnfeld, Adithya Gangidi, Adolfo Victoria, Ahuva Goldstand, Ajay Menon, Ajay Sharma, Alex Boesenberg, Alex Vaughan, Alexei Baevski, Allie Feinstein, Amanda Kallet, Amit Sangani, Anam Yunus, Andrei Lupu, Andres Alvarado, Andrew Caples, Andrew Gu, Andrew Ho, Andrew Poulton, Andrew Ryan, Ankit Ramchandani, Annie Franco, Aparajita Saraf, Arkabandhu Chowdhury, Ashley Gabriel, Ashwin Bharambe, Assaf Eisenman, Azadeh Yazdan, Beau James, Ben Maurer, Benjamin Leonhardi, Bernie Huang, Beth Loyd, Beto~De Paola, Bhargavi Paranjape, Bing Liu, Bo~Wu, Boyu Ni, Braden Hancock, Bram Wasti, Brandon Spence, Brani Stojkovic, Brian Gamido, Britt Montalvo, Carl
  Parker, Carly Burton, Catalina Mejia, Changhan Wang, Changkyu Kim, Chao Zhou, Chester Hu, Ching-Hsiang Chu, Chris Cai, Chris Tindal, Christoph Feichtenhofer, Damon Civin, Dana Beaty, Daniel Kreymer, Daniel Li, Danny Wyatt, David Adkins, David Xu, Davide Testuggine, Delia David, Devi Parikh, Diana Liskovich, Didem Foss, Dingkang Wang, Duc Le, Dustin Holland, Edward Dowling, Eissa Jamil, Elaine Montgomery, Eleonora Presani, Emily Hahn, Emily Wood, Erik Brinkman, Esteban Arcaute, Evan Dunbar, Evan Smothers, Fei Sun, Felix Kreuk, Feng Tian, Firat Ozgenel, Francesco Caggioni, Francisco Guzm\'{a}n, Frank Kanayet, Frank Seide, Gabriela~Medina Florez, Gabriella Schwarz, Gada Badeer, Georgia Swee, Gil Halpern, Govind Thattai, Grant Herman, Grigory Sizov, Guangyi, Zhang, Guna Lakshminarayanan, Hamid Shojanazeri, Han Zou, Hannah Wang, Hanwen Zha, Haroun Habeeb, Harrison Rudolph, Helen Suk, Henry Aspegren, Hunter Goldman, Ibrahim Damlaj, Igor Molybog, Igor Tufanov, Irina-Elena Veliche, Itai Gat, Jake Weissman, James
  Geboski, James Kohli, Japhet Asher, Jean-Baptiste Gaya, Jeff Marcus, Jeff Tang, Jennifer Chan, Jenny Zhen, Jeremy Reizenstein, Jeremy Teboul, Jessica Zhong, Jian Jin, Jingyi Yang, Joe Cummings, Jon Carvill, Jon Shepard, Jonathan McPhie, Jonathan Torres, Josh Ginsburg, Junjie Wang, Kai Wu, Kam~Hou U, Karan Saxena, Karthik Prasad, Kartikay Khandelwal, Katayoun Zand, Kathy Matosich, Kaushik Veeraraghavan, Kelly Michelena, Keqian Li, Kun Huang, Kunal Chawla, Kushal Lakhotia, Kyle Huang, Lailin Chen, Lakshya Garg, Lavender A, Leandro Silva, Lee Bell, Lei Zhang, Liangpeng Guo, Licheng Yu, Liron Moshkovich, Luca Wehrstedt, Madian Khabsa, Manav Avalani, Manish Bhatt, Maria Tsimpoukelli, Martynas Mankus, Matan Hasson, Matthew Lennie, Matthias Reso, Maxim Groshev, Maxim Naumov, Maya Lathi, Meghan Keneally, Michael~L. Seltzer, Michal Valko, Michelle Restrepo, Mihir Patel, Mik Vyatskov, Mikayel Samvelyan, Mike Clark, Mike Macey, Mike Wang, Miquel~Jubert Hermoso, Mo~Metanat, Mohammad Rastegari, Munish Bansal, Nandhini
  Santhanam, Natascha Parks, Natasha White, Navyata Bawa, Nayan Singhal, Nick Egebo, Nicolas Usunier, Nikolay~Pavlovich Laptev, Ning Dong, Ning Zhang, Norman Cheng, Oleg Chernoguz, Olivia Hart, Omkar Salpekar, Ozlem Kalinli, Parkin Kent, Parth Parekh, Paul Saab, Pavan Balaji, Pedro Rittner, Philip Bontrager, Pierre Roux, Piotr Dollar, Polina Zvyagina, Prashant Ratanchandani, Pritish Yuvraj, Qian Liang, Rachad Alao, Rachel Rodriguez, Rafi Ayub, Raghotham Murthy, Raghu Nayani, Rahul Mitra, Raymond Li, Rebekkah Hogan, Robin Battey, Rocky Wang, Rohan Maheswari, Russ Howes, Ruty Rinott, Sai~Jayesh Bondu, Samyak Datta, Sara Chugh, Sara Hunt, Sargun Dhillon, Sasha Sidorov, Satadru Pan, Saurabh Verma, Seiji Yamamoto, Sharadh Ramaswamy, Shaun Lindsay, Shaun Lindsay, Sheng Feng, Shenghao Lin, Shengxin~Cindy Zha, Shiva Shankar, Shuqiang Zhang, Shuqiang Zhang, Sinong Wang, Sneha Agarwal, Soji Sajuyigbe, Soumith Chintala, Stephanie Max, Stephen Chen, Steve Kehoe, Steve Satterfield, Sudarshan Govindaprasad, Sumit Gupta,
  Sungmin Cho, Sunny Virk, Suraj Subramanian, Sy~Choudhury, Sydney Goldman, Tal Remez, Tamar Glaser, Tamara Best, Thilo Kohler, Thomas Robinson, Tianhe Li, Tianjun Zhang, Tim Matthews, Timothy Chou, Tzook Shaked, Varun Vontimitta, Victoria Ajayi, Victoria Montanez, Vijai Mohan, Vinay~Satish Kumar, Vishal Mangla, V\'{\i}tor Albiero, Vlad Ionescu, Vlad Poenaru, Vlad~Tiberiu Mihailescu, Vladimir Ivanov, Wei Li, Wenchen Wang, Wenwen Jiang, Wes Bouaziz, Will Constable, Xiaocheng Tang, Xiaofang Wang, Xiaojian Wu, Xiaolan Wang, Xide Xia, Xilun Wu, Xinbo Gao, Yanjun Chen, Ye~Hu, Ye~Jia, Ye~Qi, Yenda Li, Yilin Zhang, Ying Zhang, Yossi Adi, Youngjin Nam, Yu, Wang, Yuchen Hao, Yundi Qian, Yuzi He, Zach Rait, Zachary DeVito, Zef Rosnbrick, Zhaoduo Wen, Zhenyu Yang, and Zhiwei Zhao. 2024.
\newblock \href {https://arxiv.org/abs/2407.21783} {The llama 3 herd of models}.
\newblock \emph{Preprint}, arXiv:2407.21783.

\bibitem[{Goldman et~al.(2024)Goldman, Jacovi, Slobodkin, Maimon, Dagan, and Tsarfaty}]{goldman2024really}
Omer Goldman, Alon Jacovi, Aviv Slobodkin, Aviya Maimon, Ido Dagan, and Reut Tsarfaty. 2024.
\newblock Is it really long context if all you need is retrieval? towards genuinely difficult long context nlp.
\newblock \emph{arXiv preprint arXiv:2407.00402}.

\bibitem[{Jiang et~al.(2023)Jiang, Sablayrolles, Mensch, Bamford, Chaplot, de~las Casas, Bressand, Lengyel, Lample, Saulnier, Lavaud, Lachaux, Stock, Scao, Lavril, Wang, Lacroix, and Sayed}]{jiang2023mistral}
Albert~Q. Jiang, Alexandre Sablayrolles, Arthur Mensch, Chris Bamford, Devendra~Singh Chaplot, Diego de~las Casas, Florian Bressand, Gianna Lengyel, Guillaume Lample, Lucile Saulnier, Lélio~Renard Lavaud, Marie-Anne Lachaux, Pierre Stock, Teven~Le Scao, Thibaut Lavril, Thomas Wang, Timothée Lacroix, and William~El Sayed. 2023.
\newblock \href {https://arxiv.org/abs/2310.06825} {Mistral 7b}.
\newblock \emph{Preprint}, arXiv:2310.06825.

\bibitem[{Jiang et~al.(2024)Jiang, Rajendran, Ravikumar, and Aragam}]{jiang2024llms}
Yibo Jiang, Goutham Rajendran, Pradeep Ravikumar, and Bryon Aragam. 2024.
\newblock Do llms dream of elephants (when told not to)? latent concept association and associative memory in transformers.
\newblock \emph{arXiv preprint arXiv:2406.18400}.

\bibitem[{Jin et~al.(2024)Jin, Yu, Huang, Zeng, Wang, Hua, Zhao, Mei, Meng, Ding et~al.}]{jin2024exploring}
Mingyu Jin, Qinkai Yu, Jingyuan Huang, Qingcheng Zeng, Zhenting Wang, Wenyue Hua, Haiyan Zhao, Kai Mei, Yanda Meng, Kaize Ding, et~al. 2024.
\newblock Exploring concept depth: How large language models acquire knowledge at different layers?
\newblock \emph{arXiv preprint arXiv:2404.07066}.

\bibitem[{Ju et~al.(2024)Ju, Sun, Du, Yuan, Ren, and Liu}]{ju2024large}
Tianjie Ju, Weiwei Sun, Wei Du, Xinwei Yuan, Zhaochun Ren, and Gongshen Liu. 2024.
\newblock How large language models encode context knowledge? a layer-wise probing study.
\newblock \emph{arXiv preprint arXiv:2402.16061}.

\bibitem[{Li et~al.(2024)Li, Patel, Vi{\'e}gas, Pfister, and Wattenberg}]{li2024inference}
Kenneth Li, Oam Patel, Fernanda Vi{\'e}gas, Hanspeter Pfister, and Martin Wattenberg. 2024.
\newblock Inference-time intervention: Eliciting truthful answers from a language model.
\newblock \emph{Advances in Neural Information Processing Systems}, 36.

\bibitem[{Li et~al.(2023)Li, Wang, Ma, Wu, Wang, Gao, and Liu}]{li2023split}
Zongjie Li, Chaozheng Wang, Pingchuan Ma, Daoyuan Wu, Shuai Wang, Cuiyun Gao, and Yang Liu. 2023.
\newblock \href {https://arxiv.org/abs/2310.01432} {Split and merge: Aligning position biases in large language model based evaluators}.
\newblock \emph{Preprint}, arXiv:2310.01432.

\bibitem[{Liu et~al.(2023{\natexlab{a}})Liu, Casper, Hadfield-Menell, and Andreas}]{liu2023cognitive}
Kevin Liu, Stephen Casper, Dylan Hadfield-Menell, and Jacob Andreas. 2023{\natexlab{a}}.
\newblock Cognitive dissonance: Why do language model outputs disagree with internal representations of truthfulness?
\newblock \emph{arXiv preprint arXiv:2312.03729}.

\bibitem[{Liu et~al.(2023{\natexlab{b}})Liu, Lin, Hewitt, Paranjape, Bevilacqua, Petroni, and Liang}]{liu2023lost}
Nelson~F Liu, Kevin Lin, John Hewitt, Ashwin Paranjape, Michele Bevilacqua, Fabio Petroni, and Percy Liang. 2023{\natexlab{b}}.
\newblock \href {https://arxiv.org/abs/2307.03172} {Lost in the middle: How language models use long contexts}.

\bibitem[{Marks and Tegmark(2023)}]{marks2023geometry}
Samuel Marks and Max Tegmark. 2023.
\newblock The geometry of truth: Emergent linear structure in large language model representations of true/false datasets.
\newblock \emph{arXiv preprint arXiv:2310.06824}.

\bibitem[{McInnes et~al.(2020)McInnes, Healy, and Melville}]{mcinnes2020umapuniformmanifoldapproximation}
Leland McInnes, John Healy, and James Melville. 2020.
\newblock \href {https://arxiv.org/abs/1802.03426} {Umap: Uniform manifold approximation and projection for dimension reduction}.
\newblock \emph{Preprint}, arXiv:1802.03426.

\bibitem[{nostalgebraist(2021)}]{nostalgebraist2021logitlens}
nostalgebraist. 2021.
\newblock \href {https://colab.research.google.com/drive/1MjdfK2srcerLrAJDRaJQKO0sUiZ-hQtA} {logit lens on non-gpt2 models + extensions}.

\bibitem[{Peysakhovich and Lerer(2023)}]{peysakhovich2023attention}
Alexander Peysakhovich and Adam Lerer. 2023.
\newblock \href {https://arxiv.org/abs/2310.01427} {Attention sorting combats recency bias in long context language models}.
\newblock \emph{Preprint}, arXiv:2310.01427.

\bibitem[{Shaham et~al.(2023)Shaham, Ivgi, Efrat, Berant, and Levy}]{shaham-etal-2023-zeroscrolls}
Uri Shaham, Maor Ivgi, Avia Efrat, Jonathan Berant, and Omer Levy. 2023.
\newblock \href {https://doi.org/10.18653/v1/2023.findings-emnlp.536} {{Z}ero{SCROLLS}: A zero-shot benchmark for long text understanding}.
\newblock In \emph{Findings of the Association for Computational Linguistics: EMNLP 2023}, pages 7977--7989, Singapore. Association for Computational Linguistics.

\bibitem[{Shaham et~al.(2022)Shaham, Segal, Ivgi, Efrat, Yoran, Haviv, Gupta, Xiong, Geva, Berant, and Levy}]{shaham-etal-2022-scrolls}
Uri Shaham, Elad Segal, Maor Ivgi, Avia Efrat, Ori Yoran, Adi Haviv, Ankit Gupta, Wenhan Xiong, Mor Geva, Jonathan Berant, and Omer Levy. 2022.
\newblock \href {https://aclanthology.org/2022.emnlp-main.823} {{SCROLLS}: Standardized {C}ompa{R}ison over long language sequences}.
\newblock In \emph{Proceedings of the 2022 Conference on Empirical Methods in Natural Language Processing}, pages 12007--12021, Abu Dhabi, United Arab Emirates. Association for Computational Linguistics.

\bibitem[{Shlens(2014)}]{shlens2014tutorialprincipalcomponentanalysis}
Jonathon Shlens. 2014.
\newblock \href {https://arxiv.org/abs/1404.1100} {A tutorial on principal component analysis}.
\newblock \emph{Preprint}, arXiv:1404.1100.

\bibitem[{Sky et~al.(2024)Sky, Van~Durme, Eisner, and Kedzie}]{sky2024androids}
CH-Wang Sky, Benjamin Van~Durme, Jason Eisner, and Chris Kedzie. 2024.
\newblock Do androids know they’re only dreaming of electric sheep?
\newblock In \emph{Findings of the Association for Computational Linguistics ACL 2024}, pages 4401--4420.

\bibitem[{Tang et~al.(2024)Tang, Zhang, Ma, Lin, and Ture}]{tang2024middle}
Raphael Tang, Xinyu Zhang, Xueguang Ma, Jimmy Lin, and Ferhan Ture. 2024.
\newblock \href {https://arxiv.org/abs/2310.07712} {Found in the middle: Permutation self-consistency improves listwise ranking in large language models}.
\newblock \emph{Preprint}, arXiv:2310.07712.

\bibitem[{Team et~al.(2024)Team, Mesnard, Hardin, Dadashi, Bhupatiraju, Pathak, Sifre, Rivi{\`e}re, Kale, Love et~al.}]{team2024gemma}
Gemma Team, Thomas Mesnard, Cassidy Hardin, Robert Dadashi, Surya Bhupatiraju, Shreya Pathak, Laurent Sifre, Morgane Rivi{\`e}re, Mihir~Sanjay Kale, Juliette Love, et~al. 2024.
\newblock \href {https://arxiv.org/abs/2403.08295} {Gemma: Open models based on gemini research and technology}.
\newblock \emph{arXiv preprint arXiv:2403.08295}.

\bibitem[{Templeton et~al.(2024)Templeton, Conerly, Marcus, Lindsey, and Bricke}]{scaling}
Adly Templeton, Tom Conerly, Jonathan Marcus, Jack Lindsey, and Trenton Bricke. 2024.
\newblock Scaling monosemanticity: Extracting interpretable features from claude 3 sonnet.
\newblock \url{https://transformer-circuits.pub/2024/scaling-monosemanticity/index.html}.

\bibitem[{Vaswani et~al.(2017)Vaswani, Shazeer, Parmar, Uszkoreit, Jones, Gomez, Kaiser, and Polosukhin}]{vaswani2017attention}
Ashish Vaswani, Noam Shazeer, Niki Parmar, Jakob Uszkoreit, Llion Jones, Aidan~N Gomez, {\L}ukasz Kaiser, and Illia Polosukhin. 2017.
\newblock \href {https://arxiv.org/abs/1706.03762} {{Attention is All You Need}}.
\newblock In \emph{Advances in Neural Information Processing Systems \CNFX{NeurIPS}}.

\bibitem[{Wang et~al.(2024)Wang, Ning, Pan, Wu, Guo, Deng, Bao, Wang, and Zhang}]{wang2024novelqa}
Cunxiang Wang, Ruoxi Ning, Boqi Pan, Tonghui Wu, Qipeng Guo, Cheng Deng, Guangsheng Bao, Qian Wang, and Yue Zhang. 2024.
\newblock \href {https://arxiv.org/abs/2403.12766} {Novelqa: A benchmark for long-range novel question answering}.
\newblock \emph{Preprint}, arXiv:2403.12766.

\bibitem[{Wang et~al.(2023)Wang, Cai, Chen, Liang, and Hooi}]{wang2023primacy}
Yiwei Wang, Yujun Cai, Muhao Chen, Yuxuan Liang, and Bryan Hooi. 2023.
\newblock Primacy effect of chatgpt.
\newblock \emph{arXiv preprint arXiv:2310.13206}.

\bibitem[{Zhang et~al.(2024{\natexlab{a}})Zhang, Chen, Hu, Xu, Chen, Hao, Han, Thai, Wang, Liu, and Sun}]{zhang2024inftybench}
Xinrong Zhang, Yingfa Chen, Shengding Hu, Zihang Xu, Junhao Chen, Moo~Khai Hao, Xu~Han, Zhen~Leng Thai, Shuo Wang, Zhiyuan Liu, and Maosong Sun. 2024{\natexlab{a}}.
\newblock \href {https://arxiv.org/abs/2402.13718} {$\infty$bench: Extending long context evaluation beyond 100k tokens}.
\newblock \emph{Preprint}, arXiv:2402.13718.

\bibitem[{Zhang et~al.(2024{\natexlab{b}})Zhang, Chen, Liu, Yao, Ruwase, Chen, Wu, and Wang}]{zhang2024found}
Zhenyu Zhang, Runjin Chen, Shiwei Liu, Zhewei Yao, Olatunji Ruwase, Beidi Chen, Xiaoxia Wu, and Zhangyang Wang. 2024{\natexlab{b}}.
\newblock Found in the middle: How language models use long contexts better via plug-and-play positional encoding.
\newblock \emph{arXiv preprint arXiv:2403.04797}.

\bibitem[{Zhao et~al.(2021)Zhao, Wallace, Feng, Klein, and Singh}]{zhao2021calibrate}
Zihao Zhao, Eric Wallace, Shi Feng, Dan Klein, and Sameer Singh. 2021.
\newblock \href {http://proceedings.mlr.press/v139/zhao21c/zhao21c.pdf} {Calibrate before use: Improving few-shot performance of language models}.
\newblock In \emph{International Conference on Machine Learning \CNFX{ICML}}, pages 12697--12706.

\end{thebibliography}

\newpage
\onecolumn
\appendix

\section{Prompting Details}
\label{sec:prompt}
Following setup by \citet{liu2023lost}, we construct key-value pairs retrieval and multi-document question answering prompting dataset.

\paragraph{Key-Value pairs retrieval (kv-pairs)}
We generate $n$ pairs of 128-bit randomly generated UUID. 

\begin{tcolorbox}[colframe=gray, colback=white, colframe=black!75!white, arc=0mm]
{\small \textbf{Example Key-Value pair}} \\
{\small "7f666c61-573f-4212-a0a9-6f90d487cd4a" : "2a1d0ba0-cfe4-4df5-987a-6ee1be2c6ac0"}
\end{tcolorbox}
\noindent The $n$ kv-pairs are composed into one single JSON object. To test at ID $k$, we choose one pair as gold, insert it at ID $k$, and then construct as a prompt in the format:
\begin{tcolorbox}[colframe=gray, colback=white, colframe=black!75!white, arc=0mm]
{\small Extract the value corresponding to the specified key in the JSON object below.\\\\
JSON data:\\ \{
"key$^1$: "value$^1$",\\
“key$^2$": "value$^2$",\\
...\\
\textbf{"key$^k$": "value$^k$"},\\
...\\
"key$^n$": "value$^n$",\\
\} \\\\
Key: "key$^k$"\\Corresponding value:}
\end{tcolorbox}

\paragraph{Multi-document question answering (MDQA)}
In the $n$ document setting, we randomly select one question answer pair from the dataset by \citet{liu2023lost}. Subsequently we retrieve the document containing this answer and mark it as gold.
\begin{tcolorbox}[colframe=gray, colback=white, colframe=black!75!white, arc=0mm]
{\small \textbf{Example retrieval}} \\
{\small \textbf{Question}: who got the first nobel prize in physics} \\
{\small \textbf{Answer}: Wilhelm Conrad Röntgen} \\
{\small \textbf{Document}: (Title: List of Nobel laureates in Physics) The first Nobel Prize in Physics was awarded in 1901 to Wilhelm Conrad Röntgen, of Germany, who received...}
\end{tcolorbox}
\noindent We then sample $n-1$ distractors, relevant documents that do not contain the answer. To test at ID $k$, we randomly shuffle the distractors and then insert the gold document at ID $k$. Example prompt with gold document at ID $k$ is like:
\begin{tcolorbox}[colframe=gray, colback=white, colframe=black!75!white, arc=0mm]
{\small Write a high-quality answer for the given question using only the provided search results (some of which might be irrelevant). \\

Document [1](Title: Asian Americans in science and technology) Prize in physics for discovery of the subatomic... \\
... \\
\textbf{Document [$k$](Title: List of Nobel laureates in Physics) The first Nobel Prize in Physics was awarded in 1901...} \\
... \\
Document [$n$] (Title: Scientist) and pursued through a unique method, was essentially in place. Ramón y Cajal won ... \\

Question: who got the first nobel prize in physics\\
Answer: 
}
\end{tcolorbox}






\section{Probing Setup}
\label{sec:dataset}

In the experiment described in \S\ref{sec:probing}, we employ linear classifiers as our probing method.
For any given task, we choose $\{1, 0.1n, 0.2n, \dots, 1.0n\}$-th position as gold ID, where $n$ is the number of documents ($0.1n$ for $n=30$ means at ID $3$). 
Following the prompt format in \S\ref{sec:prompt}, we generate prompts with all chosen IDs, for $10,000$ iterations, resulting in a set of 110,000 prompts.
Each prompt is fed into language model, and the embedding from each layer's last token is collected. For each layer, separately, we have $110,000$ embeddings corresponding to 11 IDs and train a classifier for ten times, with embedding as input and ID as output. We calculate their mean accuracy and standard deviation.

\section{Experiments Results on \texttt{Gemma-7b-it} \cite{team2024gemma}}
\label{sec:models}

Following \S\ref{sec:4.1} and \S\ref{sec:4:2}, we conduct same experiment procedure on one additional model, which produces the same pattern. The experiment is running on one A100 GPU.


\begin{figure}[h]
  \centering
  {
 \includegraphics[width=0.9\linewidth]{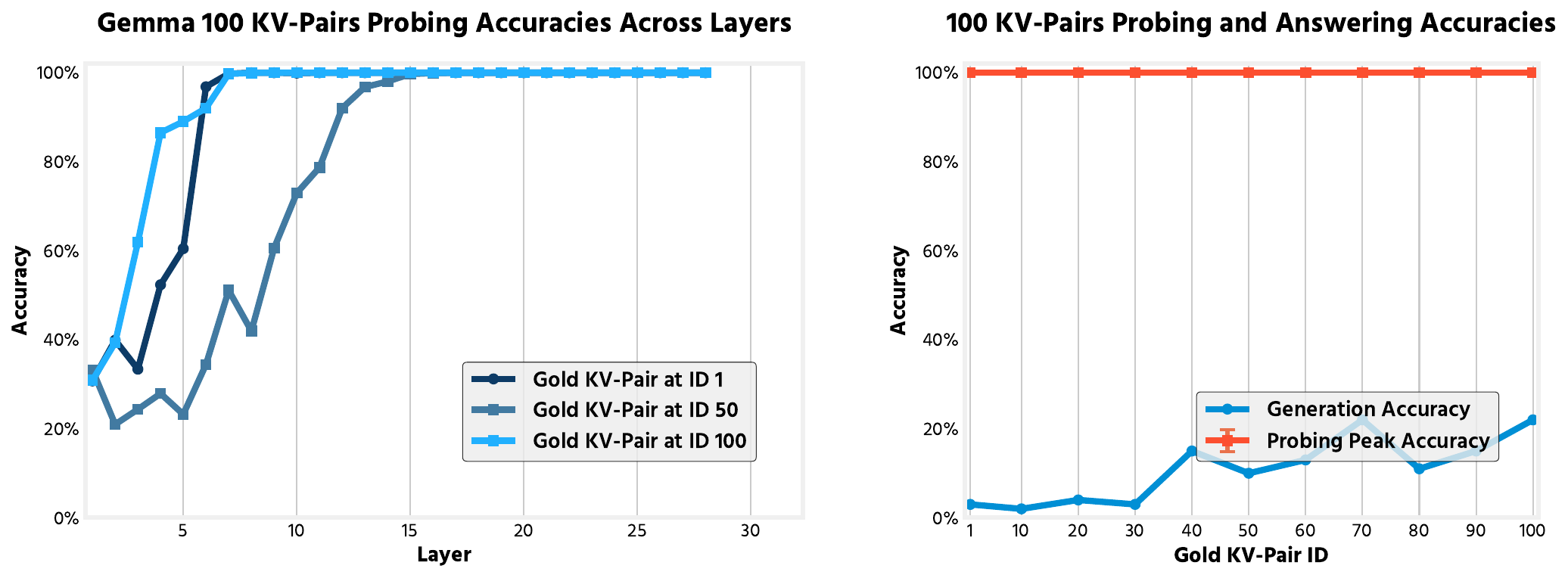}
  \caption{
  Replicating the results of Fig.\ref{fig:probing_layer_section_3} and Fig.\ref{fig:probing_comparison_section_4} using the Gemma model with \underline{100 kv-pairs}. The findings for this model also align with the observations in \S\ref{sec:4.1} and \S\ref{sec:4:2}.
  On the right, there is a notable gap between generation accuracy and peak probing accuracy, mirroring the results observed with Mistral in the main text.
  }
  }
\end{figure}
\begin{figure}[h]
  \centering
  {
 \includegraphics[width=0.9\linewidth]{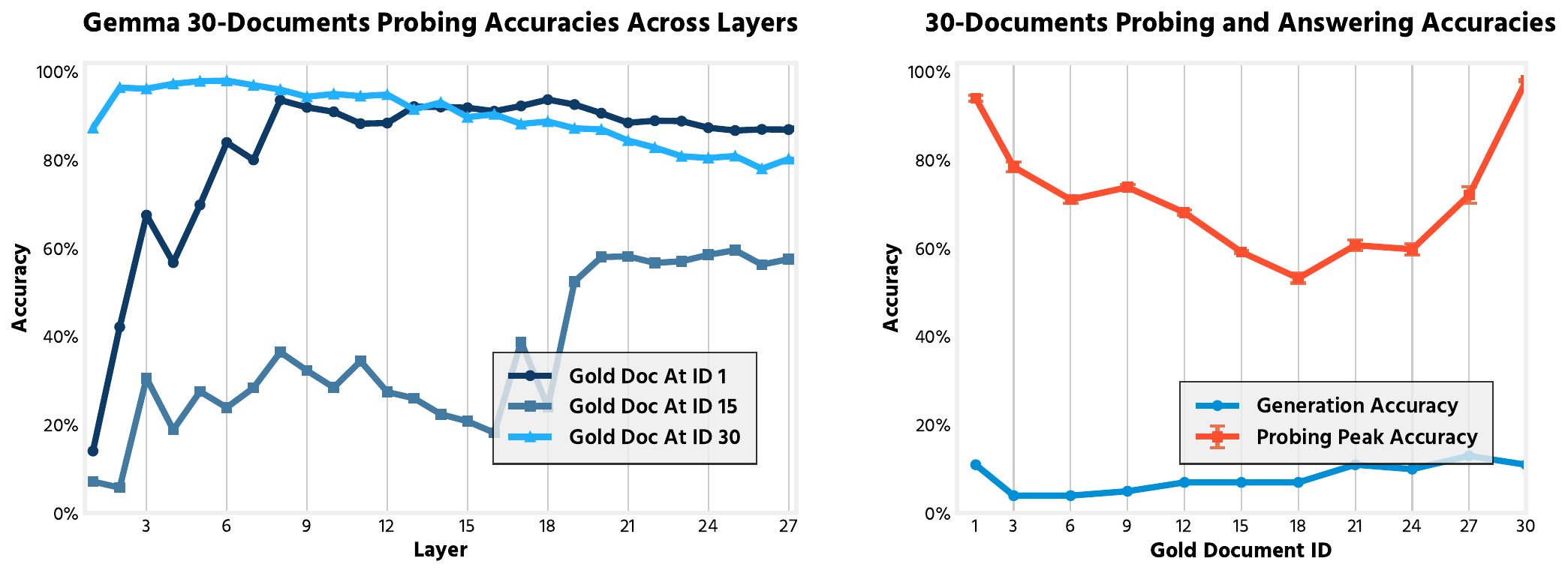}
  \caption{
  Replicating the results of Fig.\ref{fig:probing_layer_section_3} and Fig.\ref{fig:probing_comparison_section_4} using the Gemma model with \underline{30-document MDQA}. The findings for this model also align with the observations in \S\ref{sec:4.1} and \S\ref{sec:4:2}.
  On the left, while the beginning and end contexts follow the same pattern, the middle context exhibits a sudden drop in accuracy, indicating a brief loss of information, which is quickly regained.
  }
  }
\end{figure}

\clearpage
\section{2D Visualization of Hidden States Per Layer}
\label{sec:visualization}
We illustrate the model's internal representations by employing Principal Component Analysis (PCA)~\citep{shlens2014tutorialprincipalcomponentanalysis} (Fig.~\ref{fig:PCA_visualization}) and Uniform Manifold Approximation and Projection (UMAP)~\citep{mcinnes2020umapuniformmanifoldapproximation} (Fig.~\ref{fig:t-SNE_visualiation}) to reduce the hidden states to two dimensions. Each subplot depicts the hidden states of a \emph{single layer} within the Transformer. The subplots feature 11 colored points, each representing a \emph{different gold document ID}. The color-to-ID mapping can be found in the color bar located in the upper right corner.
We reuse the 30-documents prompts. Among this same set of documents, we rotate the ID of the gold document to 11 equally separated positions, as in \S\ref{sec:dataset}. For the 11 prompts, we extract the last token dimensions from \texttt{LLaMa3-8B-Instruct} and perform dimensional reduction. Each dots in the plot represent a single data point.


\begin{figure}[H]
    \centering
    \includegraphics[width=1\linewidth]{pics/PCA.pdf}
    \caption{
    PCA visualization across layers: We apply PCA dimension reduction to the last embedding of each layer and visualize the results. The color of the dots indicates the position of the gold document in the prompt, ranging from 0\% (gold document at the beginning) to 100\% (gold document at the end). We observe that the dots are entangled in the early layers, start to form a 
    path
    in the order of gold document ID in the middle layers, and this order diminishes in the later layers. This observation further supports our findings in \S\ref{sec:4:2}, indicating that the internal representation of LLMs becomes richer with more layers, although the final layer may not exhibit the peak of representation richness.
    }
    \label{fig:PCA_visualization}
\end{figure}
\newpage

\begin{figure}[H]
    \centering
    \includegraphics[width=1\linewidth]{pics/t-sne.pdf}
    \caption{UMAP visualization across layers reveals a pattern similar to that in PCA Fig.~\ref{fig:PCA_visualization}. The embeddings become less entangled in the middle layers and appear more random in the extreme (earlier and final) layers.}
    \label{fig:t-SNE_visualiation}
\end{figure}

\begin{figure}[H]
    \centering
    \includegraphics[width=0.8\linewidth]{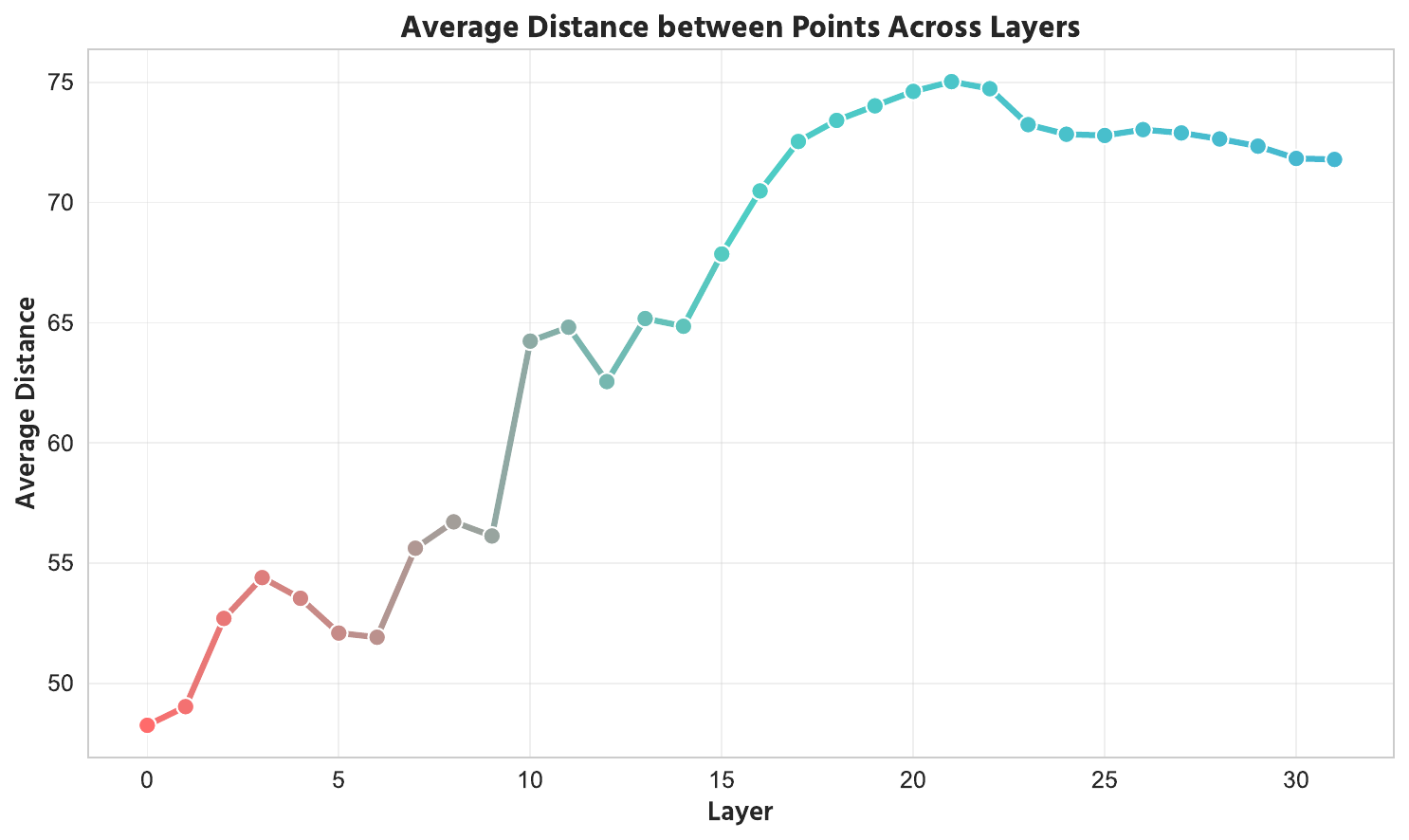}
    \caption{Average distance between scatter points across layers. The average distance has an obvious increasing trend during early layers, after achieves peak value around layer 21, it gradually decreases. The overall trend is similar to the probing accuracy in Fig.~\ref{fig:probing_comparison_section_4}.}
    \label{fig:enter-label}
\end{figure}

\newpage
Below, we calculate the average distance between points in the PCA visualization (Fig.\ref{fig:PCA_visualization}). For each layer $l$ (each subfigure), there are 11 points ${x^l_1, x^l_2, x^l_3, \dots, x^l_{11}}$, each corresponding to a different gold ID. We compute the average distance for layer $l$ as $d_l = \sum_{i=1}^{10} \text{distance}(x^l_i, x^l_{i+1}) / 10$. As shown in Fig.\ref{fig:enter-label}, the average distance $d_l$ increases up to layer 21, indicating a growing degree of separability. After reaching the peak distance, it decreases until the final layer. This pattern further supports our probing results. Additionally, as observed in \S\ref{sec:extraction timing}, most gold IDs achieve peak accuracy around layer 18, which is very close to the peak distance layer (layer 20) identified here.


\newpage
\newpage
\section{Language Generation Logits Trends Across Layers}
\label{sec:logit_lens_main}

\begin{figure}[ht]
\centering
  \includegraphics[width=0.5\columnwidth]{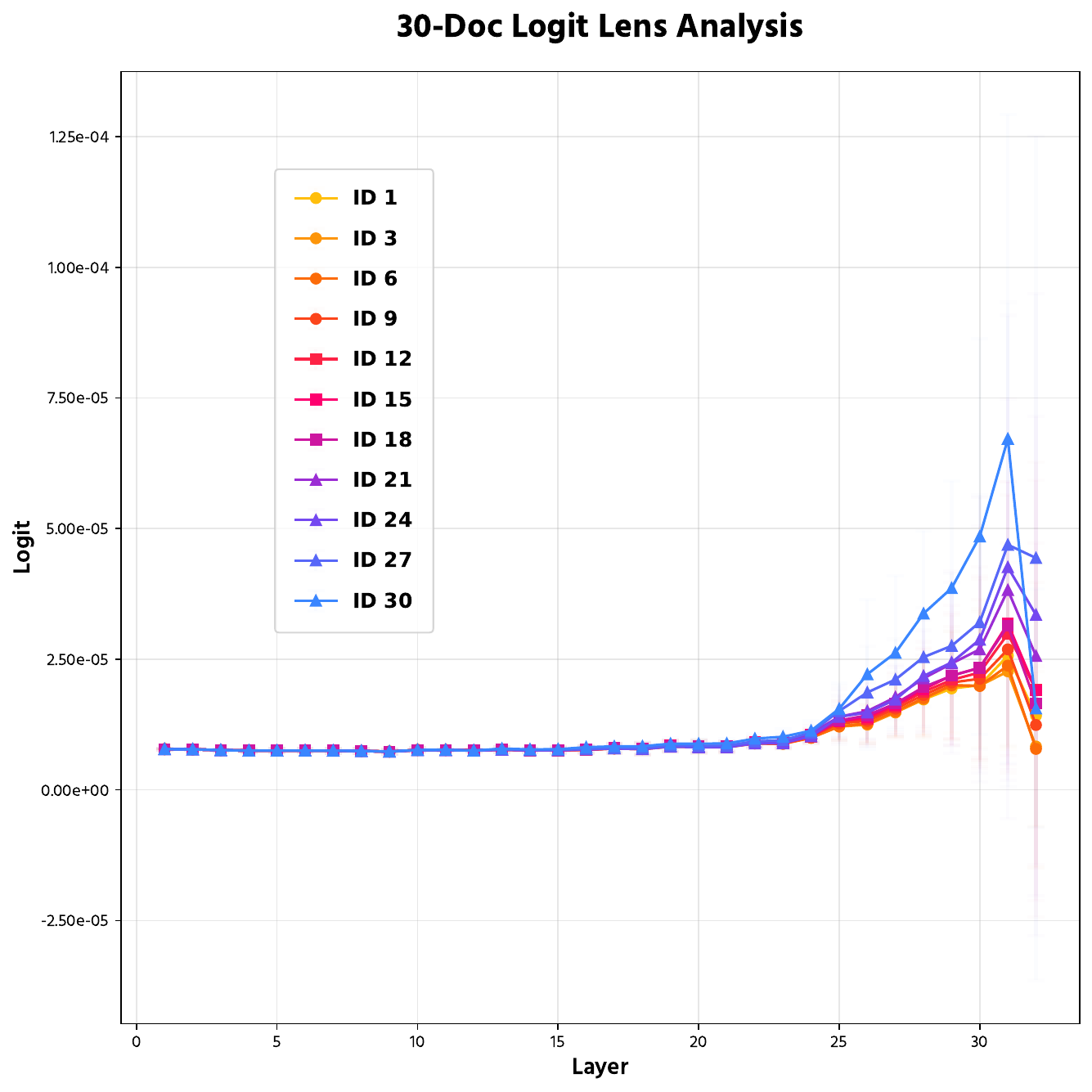}
  \caption{Logits of generating the first token of the correct answer across layers. The lines indicate different gold document ID. x-axis is the layer index, and y-axis is the logits across 1000 different prompts. During middle layers, blue line(ID 30) rises much more significantly than other IDs, however it also drops as the lowest at the last layer. For other lines, they share similar patterns in most layers but diverge in last few layers.}
  \label{fig:logit-lens}
\end{figure}

We apply the \textit{logit lens}~\citep{nostalgebraist2021logitlens} to examine the layers where the model begins to produce more probability mass over the correct answer. 
Specifically, for each layer, we multiply the last token embedding $x^l$  with the \texttt{LM Head} and apply \texttt{SoftMax} to obtain the generation distribution over the entire vocabulary. We then record the probability of generating the correct token. 
The logit lens enables internal analysis of the LLM output distribution flow across layers.
We reuse the 30-document prompts and apply the logit lens to each. Fig.~\ref{fig:logit-lens} shows how the logits for the correct answer evolve across layers, with visualizations for prompts containing the gold document at different locations.

\paragraph{Layer-wise logit divergence.}
The changes in logits across layers reveal certain key observations: 
\begin{enumerate} 
    \item[1.] In all prompts, the logits follow an almost identical pattern before layer 20, showing a steady, minimal increase. The divergence begins at layer 20. 
    \item[2.] At layer 20, when the gold document is placed at the end of the prompt, there is a sharp and significant increase in the logits. This is supported by Fig.~\ref{fig:probing_comparison_section_4}, where the same prompt achieves near-perfect probing accuracy.
    \item[3.] All prompts reach their peak logit at layer 31, but exhibit a positional bias. The later the gold document appears in the prompt, the higher the peak logit.
\end{enumerate}

These findings reveal that positional bias emerges during the model's internal processing, with a noticeable shift at a certain layer, where the position of gold information influences the quality of language output.

\end{document}